\documentclass[twoside]{article}

\usepackage[accepted]{aistats2025}

\usepackage[round]{natbib}

\usepackage[utf8]{inputenc} 
\usepackage[T1]{fontenc}    
\usepackage{hyperref}       
\usepackage{url}            
\usepackage{booktabs}       
\usepackage{amsfonts}       
\usepackage{nicefrac}       
\usepackage{microtype}      
\usepackage{xcolor}         
\usepackage{algorithm}
\usepackage[noend]{algpseudocode}
\usepackage{amsmath}
\usepackage{amssymb}
\usepackage{mathtools}
\usepackage{amsthm}
\usepackage{bm}
\usepackage{url}
\usepackage{xurl}
\usepackage{hyperref}
\usepackage{subcaption}
\usepackage{multirow}
\usepackage{array}
\usepackage{wrapfig}
\usepackage{booktabs}
\usepackage{makecell}
\usepackage{graphicx}

\usepackage{color,soul}
\usepackage{xr}
\newtheorem{theorem}{Theorem}

\newtheorem{definition}{Definition}

\def\tX{\mathbf{\tilde{X}}}
\def\x{\mathbf{x}}
\def\tx{\mathbf{\tilde{x}}}

\def\supp{\textrm{supp}}
\def\capture{\textrm{cap}}
\def\eadd{e^{\textrm{add}}}
\def\ebase{e^{\textrm{base}}}

\begin{document}

\twocolumn[

\aistatstitle{Models That Are Interpretable But Not Transparent}

\aistatsauthor{ Chudi Zhong  \And Panyu Chen \And  Cynthia Rudin }

\aistatsaddress{Duke University \\UNC-Chapel Hill \And  Duke University \And Duke University } ]

\begin{abstract}
Faithful explanations are essential for machine learning models in high-stakes applications. Inherently interpretable models are well-suited for these applications because they naturally provide faithful explanations by revealing their decision logic. However, model designers often need to keep these models proprietary to maintain their value. This creates a tension: we need models that are \textit{interpretable}—allowing human decision-makers to understand and justify predictions, 
but not \textit{transparent}, so that the model's decision boundary is not easily replicated by attackers. Shielding the model's decision boundary is particularly challenging alongside the requirement of completely faithful explanations, since such explanations reveal the true logic of the model for an entire subspace around each query point. This work provides an approach, \textit{FaithfulDefense}, that creates model explanations for logical models that are completely faithful, yet reveal as little as possible about the decision boundary. FaithfulDefense is based on a maximum set cover formulation, and we provide multiple formulations for it, taking advantage of submodularity.
\end{abstract}

\section{Introduction}
Many organizations rely on machine learning (ML) models and their employees' livelihoods rely on those models being proprietary; the competitive value of these models depends on their secrecy. At the same time, these companies would like to provide explanations for each prediction. Explanations are important for many reasons, including accountability, troubleshooting of the inputs, sanity checking, and allowing the individual subject to the scores to understand the decision and refute them \citep{rudin2019stop}. Explanations are also required by law for high stakes decisions in the European Union \citep{EuropeanParliament2016a, goodman2017european} and in the United States, via the Equal Credit Opportunity Act (ECOA). According to Rohit Chopra, the Director of the Consumer Financial Protection Bureau in the United States, ``Companies are not absolved of their legal responsibilities when they let a black-box model make lending decisions''...``The law gives every applicant the right to a specific explanation if their application for credit was denied, and that right is not diminished simply because a company uses a complex algorithm that it doesn't understand.'' \citep{cfpb_blackbox_2022}.
That is, to satisfy legal requirements, explanations must be completely \textit{faithful}, meaning that the explanations must reveal the reasoning process that is actually used in the predictive model to make the prediction. \textit{Inherently interpretable} models provide faithful explanations, but such explanations could release substantial information about the model; lenders would not use a model if it were easily reverse-engineered.

How might an organization balance between the two goals of keeping models secret and revealing information within explanations? 
This tension between keeping models interpretable and proprietary is embodied by the \textit{model extraction attack} \citep{tramer2016stealing}. The attacker queries an ML model to obtain predictions. The attacker may not know the exact model type or the true data distribution used to train the model. After enough queries, the attacker can use the collected query-label pairs to train a surrogate model that achieves accuracy very close to that of the confidential model. It is hard enough to prevent information leakage from these attacks by revealing only predictions, but the need to reveal explanations could make these attacks much more powerful. 

In this paper, we provide a model form 
that addresses this tension. We propose a novel explanation generation method for logical machine learning models, \textit{FaithfulDefense}, that ensures the released explanations are always faithful and yet does not disclose the true decision boundary used by the underlying model. Our approaches also allow organizations to customize the level of detail revealed in explanations. FaithfulDefense is derived as a maximum set cover formulation. If its solution does not use the full ``length'' budget allocated to it by the user, extra terms can be appended to preserve the disclosure of information about the model while still maintaining faithful explanations. Since maximum set cover is submodular, we also produce a greedy solution that also works well in practice. An empirical evaluation with 3 different attacker strategies and 6 different explanation methods demonstrates that our proposed method is effective in safeguarding the underlying model. An implementation of the algorithm is available at: \url{https://github.com/chudizhong/FaithfulDefense}.

\section{Related Work}

\textbf{Interpretability}: 
Understanding how predictions are made allows humans to identify and rectify potentially serious problems \citep{ashoori2019ai, brundage2020trustworthy, LoPiano20, RudinWa14, Thiebes20}.
While there is an abundance of work in 
explainable artificial intelligence (XAI), through providing simpler approximations to black-box models \citep{bastani2017interpretability, lakkaraju2019faithful} or local approximations to black boxes \citep{lundberg2020local, lundberg2017unified, ribeiro2016should, simonyan2014deep, sundararajan2017axiomatic}, such techniques are unsuitable for high-stakes decisions because their explanations are often unfaithful, incomplete or misleading \citep{lakkaraju2020fool, LaugelEtAl19, rudin2019stop, adebayo2018sanity}. Such unfaithful explanations can exacerbate problems with lack of trust. 
Interpretable machine learning, on the other hand, focuses on developing inherently interpretable models. These models reflect exactly how they make decisions, and the reasoning is always faithful. Logical models such as decision sets and decision trees are important types of interpretable models that have existed since the beginning of artificial intelligence.  
Numerous algorithms have been developed for optimizing them \citep{wang2017bayesian, aglin2020learning, angelino2018learning, DashEtAl18, demirovic2022murtree, lin2020generalized, rudin2023globally}, and modern versions of these algorithms can find sparse models with accuracy comparable to that of black box counterparts. These types of models have a long precedent in high-stakes decisions because they provide logical rules that faithfully describe, for instance, why someone's loan was denied.


\textbf{Model extraction}: Model extraction means acquiring information from an unknown target model that goes beyond simple outputs to a set of input queries.
An \textit{attacker} who extracts information from the model might aim to train its own surrogate model, having performance no worse than that of the target model \citep{tramer2016stealing,orekondy2019knockoff,shi2017steal,correia2018copycat,chandrasekaran2020exploring,shi2018active,teitelman2020stealing}. 

The notion of providing an explanation clashes with the model extraction paradigm. If an explanation is required with each prediction, the attackers job is potentially much easier since the explanation could reveal the predictions of an entire portion of the input space. 
Recent studies delve into the role of explanations in model extraction attacks \citep{yan2022towards, yan2023explanation, wang2022dualcf, ezzeddine2024knowledge, miura2024megex, oksuz2023autolycus, nguyen2023xrand}. Most of these works consider explanations derived from XAI algorithms like LIME \citep{ribeiro2016should} and SHAP \citep{lundberg2020local} or counterfactual explanations. As discussed, posthoc explanations are not faithful. They are also generally incomplete, in that they reveal possibly a few variables that might be important to the prediction, but do not reveal how these variables are used to form the prediction. Explanations from inherently interpretable models would be far more valuable to an attacker because they explain the full reasoning process that led to a decision. Given that high-stakes domains like finance generally require inherently interpretable models, and should require complete and faithful explanations, we should be far more concerned by the prospect of attackers in the setting of inherently interpretable models. 

The literature, however, focuses on the protection of individuals whose scores are computed by the model, while overlooking the need to protect the companies that develop and use these models. This gap in protection can have serious consequences: companies may avoid using inherently interpretable models out of concern for the increased risk of model extraction attacks. If companies cannot secure their proprietary interpretable models, they may continue to rely on black-box models with unfaithful explanations. 

Hence, we want models that are inherently \textit{interpretable}  -- with completely faithful explanations -- but not \textit{transparent} -- we do not want attackers to approximate or see the full model with only a few queries.


In order to gather the most valuable data to build a surrogate model, attackers can use generative models for generating artificial data for querying the target model \citep{mosafi2019stealing,kariyappa2021maze,shi2018generative,yuan2022attack,truong2021data,sanyal2022towards}, or active learning \citep{tramer2016stealing,chandrasekaran2020exploring,pal2019framework,pal2020activethief,pengcheng2018query,shi2018active,reith2019efficiently,wang2022enhance,xie2022game}.
The attacker will eventually gather enough information to build an accurate surrogate model, and our goal is to slow down this process so the model's decision boundary remains protected for as long as possible. Ideally, we want the explanations to provide little more information than if they were absent.


\section{Problem Setting}
Let $\mathcal{D} = \{(\x_i,y_i)\}_{i=1}^n$ be the dataset, where $\x_i \in \mathbb{R}^p$ are features and $y_i \in \{0, 1\}$ denotes binary labels. 
This reflects high-stakes decision-making in, for instance, loan and insurance approvals. A model $f$ can be trained to learn relations between features $\{\x_i\}$ and outcomes $\{y_i\}$. We consider logical models that use ``if-else'' conditions, specifically decision sets. Decision sets encompass other logical models, including decision trees and decision lists \citep{rudin2022interpretable}. 
Sparse decision sets are inherently interpretable and are particularly easy to understand and troubleshoot, but are also particularly difficult to protect from attackers when the true model logic is employed in explanations. Sparse generalized additive models with piecewise constant shape functions -- which are currently the most common models for financial risk scoring -- 
can be converted efficiently into decision sets, as described in Appendix \ref{ref:app:gams}. 


\begin{algorithm}
\caption{ModelExtraction($max\_q$)}\label{alg:steal_model}
    \begin{algorithmic}[1]
        \Require a maximum number of queries $max\_q$
        \State $Q \leftarrow \{\}, \textit{label} \leftarrow \{\}, E \leftarrow \{\}$ 
        \For{$t \in \{1,...,max\_q\}$}
        \State $q \leftarrow$ AttackerGenerateQuery($Q, label, E$) 
        \State $qhat, e\leftarrow$ DefenderStrategy($q$) 
        \State $Q$.add($q$), $label$.add($qhat$), $E$.add($e$)
        
        \EndFor
        \State Train a surrogate model $f'$ using $Q$, $label$ and $E$. 
    \end{algorithmic}
\end{algorithm}

An ML model extraction attack occurs when the attacker has query access to the target model $f$ and attempts to learn a surrogate model $f'$ that closely approximates or matches the performance of $f$ by sequentially asking queries. 
Conversely, the defender must provide the label of the query and give an explanation if the query has a positive label (e.g., loan denied, insurance claim denied, job application denied). 
The process is shown in Algorithm \ref{alg:steal_model}.
In this work, we stand on the position of the defender to address the challenge of how to provide meaningful and faithful explanations while preventing attackers from training an accurate surrogate model with a small number of queries.

\section{Methodology}
In this section, we first introduce our defense method, detailing how to generate an explanation for a query predicted to be positive by the model $f$. We then illustrate how attackers leverage these explanations to launch more potent attacks.

\subsection{How a Defender Can Generate Explanations}\label{sec:defender}

We assume $f$ is trained on proprietary dataset $\mathcal{D}$ and belongs to an inherently interpretable model class, such as decision sets. A decision set, also known as a ``disjunction of conjunctions,'' ``disjunctive normal form'' (DNF), or an ``OR of ANDs'' is comprised of an unordered collection of rules, where each rule is a conjunction of conditions. A positive prediction is made if at least one of the rules is satisfied. Decision sets encompass decision trees and decision lists: any decision tree or decision list can be written as a decision set. A decision set is inherently interpretable and the satisfied rule can be directly used as a faithful explanation. However, generating explanations in this way discloses too much information. We now introduce a more frugal method to provide faithful explanations. 

Given a dataset $\mathcal{D} = \{\x_i, y_i\}_{i=1}^n$, we can turn it into a binary dataset by considering all categories for categorical features and binning continuous features using thresholds. Let $C=\{c_1, c_2, ..., c_m\}$ denote the full set of possible binary features. It can also be viewed as a collection of conditions. For example, a $c_j$ could be ``income $<$5K'', ``age $\leq 20$'', ``saving $<$ 5K'', etc. The binarized dataset is denoted as $\tX=\{\tx_i\}_{i=1}^n$ where $\tx_i \in \{0, 1\}^m$ is 
a binary vector of length $m$. 
Each rule in $f$ is a subset of $C$. 
Let $\beta$ be a set of conditions. We define $\capture(\beta, \tx_i) = 1$ if $\tx_i$ satisfies all conditions in $\beta$, and 0 otherwise. The collection of samples from the dataset captured by $\beta$ can be denoted as $S_{\beta}(\tX) = \{\tx_i: \capture(\beta, \tx_i) = 1\}$. 


Let us define a faithful explanation for query $q$.
\begin{definition}
(Faithful explanation $e$ for query $q$): Let $q$ be a query such that its predicted label $f(q)$ is 1, where an explanation is required. Denote a rule in $f$ that $q$ satisfies as $f_r$, where $f_r\subseteq C$. Then, $e \subseteq C$ is a faithful explanation for $q$ if $f_r \subseteq e$.
\end{definition}

Intuitively, an explanation is faithful if it reveals part of the model. The explanation does not need to reveal \textit{all} rules in the model, or even an entire rule, to be faithful.
For instance, if a rule in the model is ``income $<$5K,'' then an explanation that ``income $<$5K and age $\leq 20$'' is faithful, and yet does not reveal that the entire ``income $<$5K'' subgroup has label 1. Providing an explanation that reveals only part of the model's original rule, rather than the full rule (``income $<$5K and age $\leq 20$'' instead of ``income $<$5K'') is identical to the process of (1) creating an equivalent underlying model with the rule split in two (i.e., replacing ``income $<$5K'' in the model with the pair of rules  ``income $<$5K and age $\leq 20$,'' ``income $<$5K and age $> 20$''), and then (2) providing a rule in this new model as the explanation (``income $<$5K and age $\leq 20$'').
Thus, our definition of a faithful explanation takes into account this option for the modeler of creating an equivalent decision set before providing the explanation. Importantly, a faithful explanation does not need to be ``complete,'' meaning it need only reveal one reason for the decision, not all possible reasons. E.g., both ``income $<$5K'' and ``employed = no'' are true, but we need only give an explanation involving one of them. (The reason would come with a notification that there may be other reasons why the loan was denied, to prevent a misinterpretation that the explanation is complete.)

We define $|e|$ as the size of the explanation, i.e., the number of conditions in the explanation. 
Let $\supp(e, \tX)$ denote the number of samples captured by $e$, i.e., 
$$\supp(e, \tX) = \sum_{i=1}^n \capture(e, \tx_i) = |S_{e}(\tX)|.$$
$\supp(e, \tX)$ can be used to estimate the released information. There is a tradeoff between $\supp(e, \tX)$ and $|e|$. Typically, $\supp(e, \tX)$ monotonically decreases as $|e|$ increases, e.g., the size 2 rule ``income $<$5K and age$\leq$ 20'' has less support than the size 1 rule ``income $<$5K.'' 

Our goal is to find a simple explanation $e$, meaning the explanation can be described succinctly, i.e., $|e|$ is no larger than a max length $l$, and yet captures few positively predicted samples in the training set. Therefore, the optimization problem can be written in the following form: 
\begin{equation}\label{eq:prob}
\begin{aligned}
    \min_{\textrm{faithful explanation  } e}   \supp(e, \tX) \ 
    \textrm{s.t.} \ \  |e| \leq l.
\end{aligned}
\end{equation}
Since $f$ is a decision set, we can slightly modify Problem \eqref{eq:prob}. 
Let $e$ consist of two parts, i.e., $e = \{\ebase, \eadd\}$, where $\ebase = f_r$ is a set of conditions used by one of the rules in $f$ that $q$ satisfies, and $\eadd$ is the set of additional conditions satisfied by $q$ other than the conditions used by the satisfied rule, denoted as $C_{q} \subseteq C\backslash \ebase$. E.g., if $q$ obeys ``income $<10k$,'' and income is not one of the conditions in $\ebase$, we may choose to add it to $\eadd$ to narrow the explanation. Since $\ebase$ is fixed, we only need to optimize $\eadd$. Then, the modified optimization problem is 
\begin{equation}\label{eq:prob2}
    \min_{e^{\textrm{add}} \subseteq C_{q}} \ \ \supp(\{e^{\textrm{base}}, e^{\textrm{add}}\}, \tX) \quad \textrm{s.t.} \ \ |e^{\textrm{add}}| \leq l.
\end{equation}

\begin{theorem}
    Let $q$ be the query with $f(q)=1$, $e^{\textrm{base}}$ be the set of conditions used by the rule in $f$ that $q$ satisfies, and $C_{q} \subseteq C \backslash e^{\textrm{base}}$ be the additional conditions satisfied by $q$. Problem \ref{eq:prob2} of selecting a subset  $e^{\textrm{add}} \subseteq C_{q}$ such that the intersection of the selected set of samples covers the minimum number of elements in $\tX$ is equivalent to the maximum coverage problem: selecting a subset $e^{\textrm{add}} \subseteq C_{q}$ such that the union of selected samples cover the maximum of the complement set $\tX^c$.
\end{theorem}

This theorem indicates that Problem \ref{eq:prob2} is the same as the maximum coverage problem, and we need only optimize $e^{\textrm{add}}$, the set of additional conditions beyond the base model to include within the explanation, though this is also intuitive.




The maximum coverage problem is NP-hard \citep{karp2010reducibility} but a submodular problem \citep{nemhauser1978analysis}, and we use three methods 
to find $e$ for each query $q$. 

\textbf{Method I: FaithfulDefense Greedy}. We choose the condition $c_j$ in each step such that $S_{\neg c_j}(\tX_{\textrm{base}})$ is maximized. The method approximates the optimal solution in a factor of $1-\frac{1}{e}$ \citep{nemhauser1978analysis}. 
If the greedy solution does not use all the length budget, i.e., $|\eadd| < l$, we randomly append extra conditions to use the full budget. 

\textbf{Method II: FaithfulDefense IP}. We can find an optimal solution for Problem \eqref{eq:prob2} by solving an integer programming (IP) problem. The IP formulation is shown below. Let $\mathbf{u} \in \{0,1\}^{n}, \mathbf{v} \in \{0,1\}^m$ be the binary vectors. Despite IP being NP-hard, the problem is usually solved in seconds by a solver.
\begin{eqnarray}
    \max_{\mathbf{u}, \mathbf{v}} && \sum_{\{i: \tx_i \textrm{not covered}\}}^n u_i \label{eq:mip_obj}\\
    s.t. && \sum_{j=1}^m v_j \leq l, \quad  \sum_{j:x_{ij} = 1} v_j \geq u_i \\
    && u_i \in \{0,1\}\  \forall i\in\{1,...,n\} \label{eq:mip_u}\\
    && v_j \in \{0,1\} \ \forall j\in \{1,...,m\}\label{eq:mip_v}
\end{eqnarray}

Equation \ref{eq:mip_obj} aims to maximize the sum of the covered samples. Equation \ref{eq:mip_u}-\ref{eq:mip_v} indicate that if $u_i=1$, $\tx_i$ is covered by at least one of the $\neg c_j$'s and if $v_j=1$, $\neg c_j$ is selected), respectively. 

\textbf{Method III: FaithfulDefense IP-RA}. If the IP solution does not use the full length budget, we randomly append extra conditions to use the entire budget. 


Algorithm \ref{alg:safeguard} shows how the defender safeguards the model while providing a faithful explanation. 

\begin{algorithm}[h]
\caption{FaithfulDefense($q$)}\label{alg:safeguard}
    \begin{algorithmic}[1]
        \Require Base model $f$, dataset $\tX$, conditions $C$, query $q$, history of explanations $E$
        \If {$f(q) = 0$} 
        \State return $f(q), \emptyset$ \Comment{No need for explanation.}
        \Else
        \If {$q$ can be explained by an  $e\in E$}
        \State $e \leftarrow$ FindFromHistory($E, q$) \Comment{Return only the explanation shown previously.}
        \Else 
        \State $e \leftarrow \textrm{GenerateExplanation}(q, f, \tX, C)$. \Comment{Solve Problem \ref{eq:prob2} by IP, IP-RA or greedy method.}
        \EndIf
        \State return $f(q), e$ \Comment{Returns low support and small size explanations}
        \EndIf
    \end{algorithmic}
\end{algorithm}



\subsection{How an Attacker Can Use Explanations}\label{sec:perturbation}
Given explanations in addition to query labels, attackers will use these explanations to generate more informative queries. A straightforward attacker strategy is generating an unasked query that lies beyond the boundaries marked by past explanations. This strategy does not fully take advantage of these explanations. An alternative strategy is to explore a high-density area close to the decision boundary. Let us explain it in detail. 

Given a query $q$ and its explanation $e$, the attacker can generate candidate queries by adjusting one of the feature values close to and outside the boundary marked by $e$. For example, if $e$ involves $k$ features, the attacker can generate at least $k$ candidate queries. However, this could lead to too many candidate queries if $|e|$ is large. Given the limited number of queries the attacker can ask, it would be reasonable to only consider candidate queries that explore regions determined by more important features. So the next question is how to determine which features are more important. The attacker can track the number of times each feature appears in past explanations $E$. Features with higher counts can be viewed as more important than others. Then the attacker can select the top $k$ features in $e$ and generate new queries that differ from $q$ by adjusting the value of one of the selected features. If a feature $j$ used in $e$ appears most often in past explanations, the new value assigned to feature $j$ is positioned close to, yet beyond, the boundary provided by $e$. For categorical features, $q_{j}^{new}$ is set to $q_{j} \pm 1$ to transition to another category. For continuous features, $q_{j}^{new}$ is adjusted by $\text{boundary}(e, j) \pm \delta$. For instance, if the explanation for feature $j$ is ``$j \leq \$ 5000$,'' then the new value for $j$ could be set to $5001$. In cases where feature $j$ is bounded by $e$ on both sides, the attacker can generate two candidate queries. All other features retain the same values as $q$. Thus, a new candidate query is formulated as $q^{new} = [q_{1}, ..., q_{j-1}, q_{j}^{new}, q_{j+1}, ...q_{p}]$. The attacker then adds all new candidate queries into a query pool and rearranges them based on the importance of the perturbed feature.

This perturbation-based query generation method is inspired by \citep{oksuz2023autolycus}, which adopts the perspective of attackers upon receiving LIME explanations \citep{ribeiro2016should}. In their setup, a LIME explanation is provided for each query, reporting weights for each feature. They then slightly adjust the value of one of the features with a high weight as a candidate query. However, our explanations differ from LIME in three ways: (1) our explanations are always faithful, following the rules used to make predictions, while LIME explanations are local approximations of the underlying model; (2) our explanations do not assign weights to features since features are not weighted in the ground truth model, so the explanation itself cannot be used to rank features; and (3) our explanations are provided only for queries predicted to be positive, while in \citep{oksuz2023autolycus}, a LIME explanation is provided for each query. 

Meanwhile, since our explanations are faithful, they can also be used as part of the attacker's surrogate model for prediction. For instance, upon the arrival of a new sample, the attacker initially verifies whether the sample aligns with any of the explanations. If such a match is found, a positive prediction is made directly. Otherwise, a prediction is generated based on the surrogate model $f'$, i.e., 
\begin{equation}\label{eq:pred}
    \hat{y}_i = \capture(E, \x_i) \vee f'(\x_i).
\end{equation}

\begin{figure*}[ht]
    \centering
    \includegraphics[width=.7\textwidth]{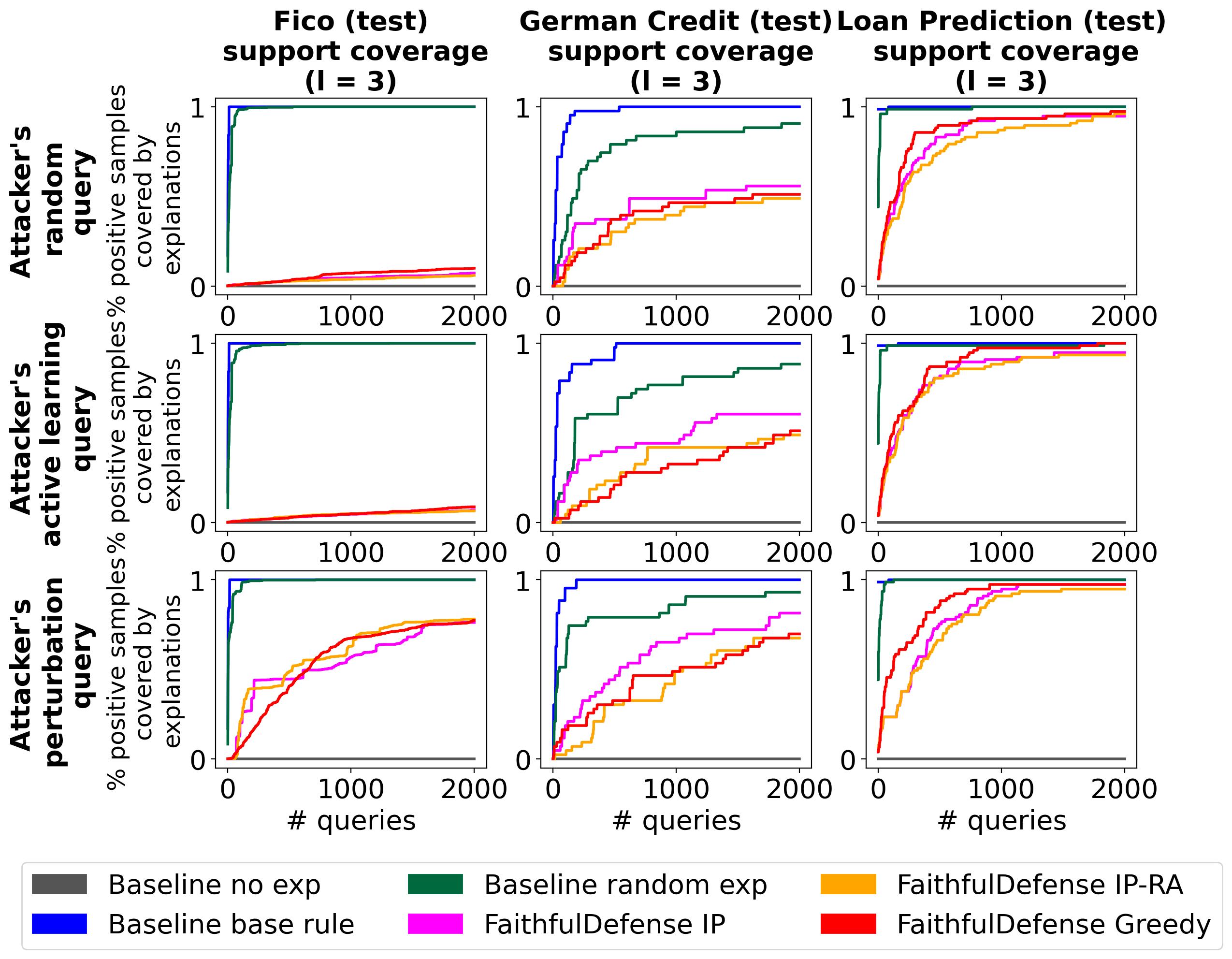}
    \caption{Number of queries vs$.$ the proportion of positive samples covered by explanations. Lower curves are better. FaithfulDefense (red, orange, pink curves) captures fewer positive samples in test sets for all three datasets using three different querying strategies. LIME was only used for perturbation queries since attackers perturb the LIME explanations to generate the next query; LIME explanations are only applicable to the attacker's perturbation-based querying strategy. LIME explanations are incompatible with other attacker querying strategies. (max length $l=3$)}
    \label{fig:supp}
\end{figure*}

\section{Experiments}
In this section, we conduct experiments to show that (1) FaithfulDefense is efficient in that it reveals less information about the dataset than baselines (Section \ref{sec:info_leak}), (2) FaithfulDefense can provide explanations within seconds (Section \ref{sec:time}), (3) FaithfulDefense requires more queries by the attacker to achieve performance similar to $f$ on the test set (Section \ref{sec:test_perform}), and (4) FaithfulDefense always provides completely faithful explanations (Section \ref{sec:faithfulness}). 

Since loan decisions are a key situation where faithful explanations are required and where model owners keep models secret,  we use three credit datasets, the Fair Isaac (FICO) credit risk dataset \citep{competition} used for the Explainable ML Challenge and German credit dataset from UCI \citep{Dua:2019}. We use the merged training/testing sets of a loan approval prediction problem dataset from Kaggle \citep{kaggle}.  Each dataset is divided into train and test sets with an 80:20 split. Details are in Appendix \ref{app:setup}. We use fast sparse rule sets (FastSRS) \citep{LiFastSRS} on the training set to train the underlying interpretable model that we aim to protect. 
       

Since, as discussed in the related work section, we know of no previous method that provides fully faithful explanations that is resilient to model extraction attacks, we compare FaithfulDefense with three natural baselines:  (1) returning the rule matched by the query in model $f$ (2) randomly appending $l$ conditions, (3) providing no explanation. We also use LIME explanations \citep{ribeiro2016should} as another baseline. 
We were able to coerce LIME into comparison because its explanations share a format (e.g., if x1>5 and x2<3, predict yes) similar to ours. Nevertheless, it's important to notice that LIME does not faithfully provide the reasoning process of the underlying model. 

We cannot include Shapley value-based explanation methods as a baseline in our comparison because they 
return only feature importances: Shapley values provide \textit{no} explanation of the underlying model's calculations, only what variables are estimated to be important in those calculations; they provide \textit{no} explanations about any observation other than the one being considered (e.g., ``the loan was denied because some of your features are more important than others''). To be faithful, the user would need to know \textit{how} the variables are combined, not just how important they might be.

We assume the attacker knows the marginal distribution of each feature but not the joint distribution. 
There are three query generation strategies that are reasonable for the attacker: (1)\label{random_query} randomly generate an unasked query outside the boundaries marked by past explanations, (2) use an agnostic active learning algorithm called importance weighted active learning (iwal) \citep{beygelzimer2010agnostic, chandrasekaran2020exploring} to find an informative query that is outside the explanation boundaries, and (3) generate queries by adjusting the value of important features as described in Section \ref{sec:perturbation}. For each dataset, we allow the attacker to ask 2000 queries. Note that only perturbation-based queries are generated if LIME is used to provide explanations. We use the absolute value of coefficients in LIME explanations as the measure of variable importance.

\subsection{How much information do the explanations leak?}\label{sec:info_leak}

Figure \ref{fig:supp} demonstrates that \textit{FaithfulDefense} (red, orange, and pink curves) \textbf{reveals less information about the test set than the baseline defense strategies} (i.e., captures fewer positive samples) for all three datasets using three different querying strategies, as the three curves consistently fall below the green and blue curves. This figure also indicates that the perturbation-based querying method is usually more aggressive than the other two attacker methods. Training results are in Appendix \ref{app:more_results}. 



\subsection{How much time does each method take to generate explanations?}\label{sec:time}
\begin{figure}[ht]
\centering
\includegraphics[width=0.49\textwidth]{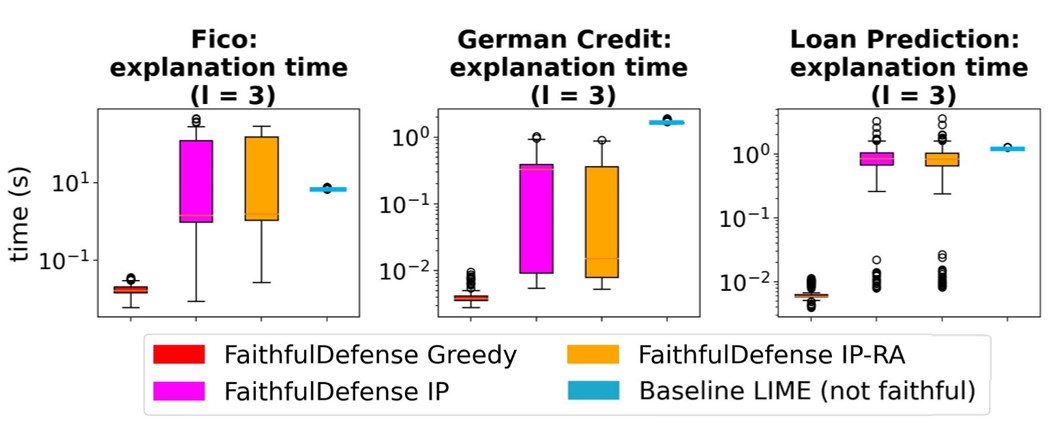}
\caption{Time consumption of generating explanations when the attacker uses the perturbation strategy.(max length $l=3$).}
\label{fig:exp_time}
\end{figure}

Figure \ref{fig:exp_time} shows the difference in time consumption between our FaithfulDefense methods and LIME for generating explanations.
\textbf{The time taken by our FaithfulDefense Greedy is generally very fast, usually less than 0.1 second.} In most cases, FaithfulDefense IP and FaithfulDefense IP-RA take longer than the greedy method but are faster than LIME.  Sometimes they can take longer because a solver is used to find the optimal solution. In practice, if the IP method cannot easily find the optimal solution, it would be better to use FaithfulDefense Greedy instead. Returning base rules as explanations and using random explanations are instantaneous. More time consumption results are in Appendix \ref{app:more_results}.

\subsection{How well does the attacker's surrogate model perform?}\label{sec:test_perform}

The attacker can train a surrogate model using the collected query-label pairs. In real situations, the attacker may not know the exact model class and architecture of the underlying model. However, they may infer that the underlying model is a logical model based on the explanations provided. Therefore, we consider three different model classes (CART, \citealt{breiman1984classification}, Random Forest, \citealt{breiman2001random}, and Gradient Boosted Tree, \citealt{friedman2001greedy}) that the attacker might use to train a surrogate model $f'$. Since all explanations except LIME are faithful, they can also be used for prediction in addition to generating more informative queries. Given a test set, Eq \eqref{eq:pred} is used to make predictions. We compare the performance based on the similarity between predictions made by the surrogate model and the underlying model on the test set, i.e., $\frac{1}{n_{\textrm{test}}}\sum_{i=1}^{n_{\textrm{test}}} 1[f(\x^{\textrm{test}}_i) == \capture(E, \x^{\textrm{test}}_i) \vee f'(\x^{\textrm{test}}_i)]$. A new surrogate model is trained on the cumulative set once 50 more query-label pairs are obtained. This allows us to track how many queries are needed to achieve a similar performance as the underlying model.

\begin{figure*}[ht]
    \centering
    \includegraphics[width=0.8\linewidth]{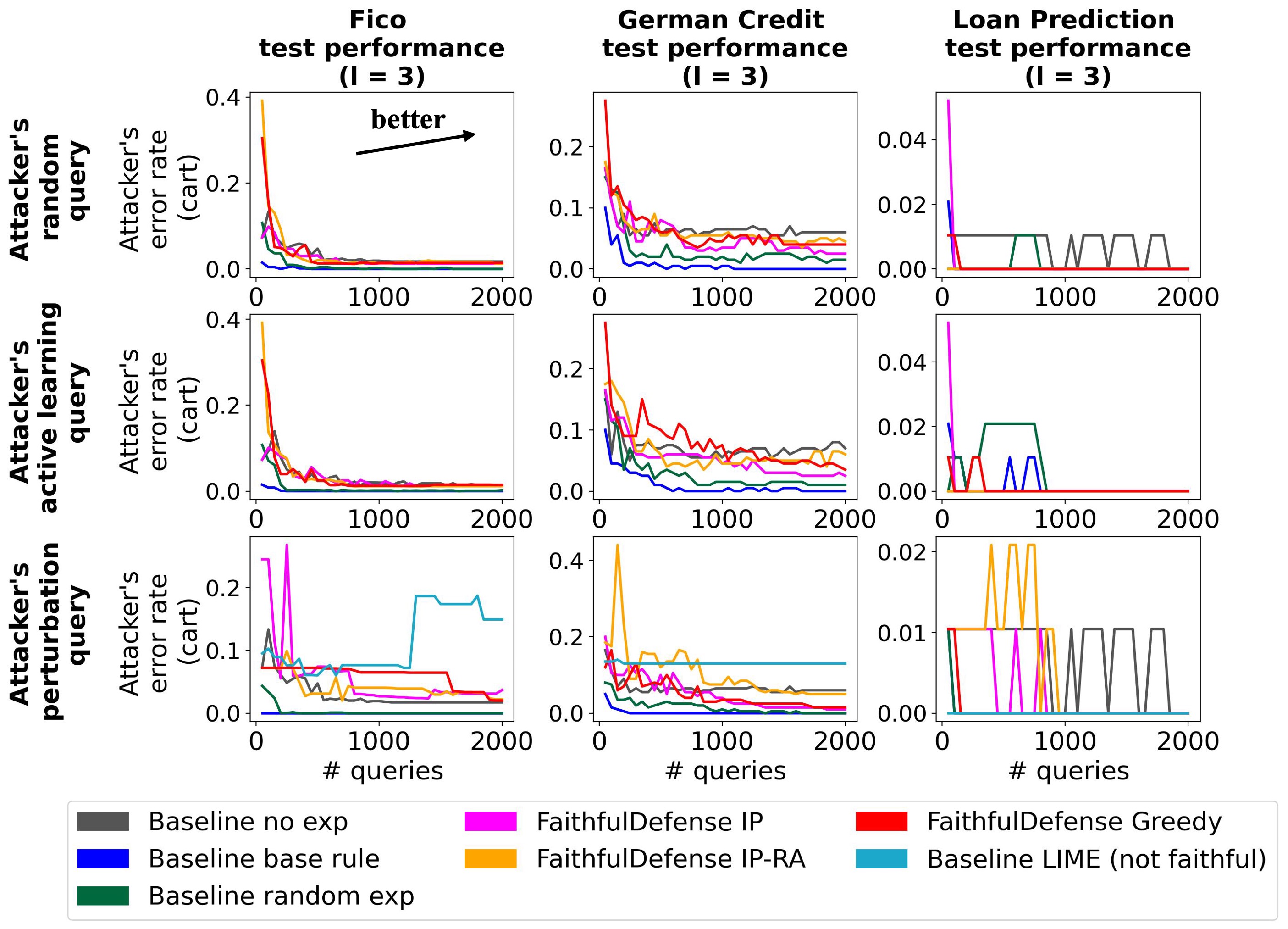}
    \caption{Comparison of test performance between base and surrogate models. CART is used by the attacker to train the surrogate model. (max length $l=3$).}
    \label{fig:cart_surrogate}
\end{figure*}

Figure \ref{fig:cart_surrogate} compares the test performance among different explanation methods. CART with a depth of 5 is used to train the surrogate model. In this figure, a higher attacker's error indicates better defense. In other words, more queries are needed to achieve similar performance to the underlying model $f$. As observed, our method (red, orange, and pink curves) is usually above the green and blue curves on the FICO and German credit datasets even when 2000 query-label pairs are used, indicating that \textbf{our explanations are more effective in protecting the underlying model}. The phenomenon is less obvious in the loan prediction dataset. The results imply that our defense methods outperform the baselines more significantly in larger datasets. 
Sometimes, we even observe overlap between the red, orange, and pink curves and the dark gray curve, suggesting that \textbf{providing explanations using FaithfulDefense is nearly equivalent to providing no explanations}. 

The bottom row in Figure \ref{fig:cart_surrogate} shows that sometimes when the attacker uses the perturbation-based querying strategy, FaithfulDefense is better than or similar to LIME in the 1000 queries and then becomes slightly worse. This is because explanations given by our FaithfulDefense are always faithful (see Table \ref{tab:fpr}) and can be used as part of the surrogate model for prediction but LIME explanations are permitted to be unfaithful. 
More results are in Appendix \ref{app:more_results}.

\subsection{How faithful the explanations are?}\label{sec:faithfulness}


We use the false positive rate (FPR) as a measure of the faithfulness of our explanations because an explanation is considered faithful when it aligns with the (partial) true model. In our framework, explanations are provided specifically when the true model predicts a positive outcome. Consequently, for samples that match the explanation, we expect them to consistently receive a positive prediction. A non-zero FPR would indicate a misalignment between the explanation and the true model. 
Table \ref{tab:fpr} shows the false positive rate on the test set when explanations themselves are used to make predictions. \textbf{All explanation methods except LIME have completely faithful explanations} as their FPR are always 0, while LIME explanations are not faithful given FPR greater than 0. 

\begin{table}[h]
\caption{False positive rate (FPR) on the test set when only explanations are used to make predictions. All explanation methods except LIME have completely faithful explanations, so their FPRs are zero.}
\centering
\begin{tabular}{|m{3.6cm}|m{0.95cm}|m{1.1cm}|m{0.85cm}|}
  \hline
  \textbf{FPR} 
   & FICO & German Credit & Loan \\\hline
  Base rule & 0 & 0 & 0 \\\hline
  Random exp & 0 & 0 & 0 \\\hline
  FaithfulDefense IP & 0 & 0 & 0 \\\hline
  FaithfulDefense IP\_RA & 0 & 0 & 0 \\\hline
  FaithfulDefense Greedy & 0 & 0 & 0 \\\hline
  LIME & 19.78\% & 15.71\% & 3.33\%\\\hline
\end{tabular}
\label{tab:fpr}
\end{table}




\section{Discussion}
Our goal is to provide models that are \textit{interpretable}, meaning that they have completely faithful explanations and could be used in high-stakes situations, but not \textit{transparent}, meaning that they provide some level of protection for the model's formula, which enables organizations to protect their efforts and intellectual property. There is no perfect protection from a large attack, in the sense that even without the requirement of explanations, an attacker could create a surrogate model based on queries alone, given enough of them. Explanations make this problem worse in that, unless they are extremely specific to the query (``people with exactly your credit history all had their loan applications denied'') they tend to reveal whole subspaces. In this work, we aimed to strike a balance between providing faithful explanations and protecting models.  

Our approach can be directly used in practice for an immense number of applications, and would be valuable to those creating models for credit scoring, car insurance, health insurance, advertising or news agencies that may reject ads or articles based on content restrictions, social media companies that can reject posts, employment, rental housing applications, and utilities (water/natural gas/electricity). 

Our faithful explanations can benefit users in several ways: they can verify their data to ensure the decision is accurate, appeal the decision if the model's logic appears to be faulty, consider legal action against the decision-maker, make improvements to their situation, or reach out to decision-makers for further information.

We believe this work may open an interesting exchange within the research community about how much information should be shared with end-users regarding the explanation behind their decisions from the lens of protecting intellectual property.



\bibliographystyle{plainnat} 
\bibliography{biblio}


\section*{Checklist}



 \begin{enumerate}

 \item For all models and algorithms presented, check if you include:
 \begin{enumerate}
   \item A clear description of the mathematical setting, assumptions, algorithm, and/or model. [Yes] Please see Sections 4 and 5.
   \item An analysis of the properties and complexity (time, space, sample size) of any algorithm. [Yes] Please see Section 4.
   \item (Optional) Anonymized source code, with specification of all dependencies, including external libraries. [Yes] We will provide the code in the supplement. 
 \end{enumerate}

 \item For any theoretical claim, check if you include:
 \begin{enumerate}
   \item Statements of the full set of assumptions of all theoretical results. [Yes] Please see Section 4. 
   \item Complete proofs of all theoretical results. [Yes] Please see Appendix. 
   \item Clear explanations of any assumptions. [Yes]     
 \end{enumerate}

 \item For all figures and tables that present empirical results, check if you include:
 \begin{enumerate}
   \item The code, data, and instructions needed to reproduce the main experimental results (either in the supplemental material or as a URL). [Yes] A summary of datasets and experimental setup are discussed in the Appendix. We will provide the code in the supplement.
   \item All the training details (e.g., data splits, hyperparameters, how they were chosen). [Yes] Please see Appendix. 
         \item A clear definition of the specific measure or statistics and error bars (e.g., with respect to the random seed after running experiments multiple times). [Yes] Please see Section 5 and Appendix. 
         \item A description of the computing infrastructure used. (e.g., type of GPUs, internal cluster, or cloud provider). [Yes] Please see Appendix. 
 \end{enumerate}

 \item If you are using existing assets (e.g., code, data, models) or curating/releasing new assets, check if you include:
 \begin{enumerate}
   \item Citations of the creator If your work uses existing assets. [Yes]
   \item The license information of the assets, if applicable. [Not Applicable]
   \item New assets either in the supplemental material or as a URL, if applicable. [Not Applicable]
   \item Information about consent from data providers/curators. [Not Applicable]
   \item Discussion of sensible content if applicable, e.g., personally identifiable information or offensive content. [Not Applicable]
   All datasets and baselines we use are publicly available. 
 \end{enumerate}

 \item If you used crowdsourcing or conducted research with human subjects, check if you include:
 \begin{enumerate}
   \item The full text of instructions given to participants and screenshots. [Not Applicable]
   \item Descriptions of potential participant risks, with links to Institutional Review Board (IRB) approvals if applicable. [Not Applicable]
   \item The estimated hourly wage paid to participants and the total amount spent on participant compensation. [Not Applicable]
 \end{enumerate}

 \end{enumerate}

%

%

\onecolumn
\appendix
\aistatstitle{Models That Are Interpretable But Not Transparent\\
Supplementary Materials}

\section{Theorems and proofs}

\textbf{Theorem 1.}
\textit{Let $q$ be the query with $f(q)=1$, $e^{\textrm{base}}$ be the set of conditions used by the rule in $f$ that $q$ satisfies, and $C_{q} \subseteq C \backslash e^{\textrm{base}}$ be the additional conditions satisfied by $q$. Problem \ref{eq:prob2} of selecting a subset  $e^{\textrm{add}} \subseteq C_{q}$ such that the intersection of the selected set of samples covers the minimum number of elements in $\tX$ is equivalent to the maximum coverage problem: selecting a subset $e^{\textrm{add}} \subseteq C_{q}$ such that the union of selected samples cover the maximum of the complement set $\tX^c$.}

\begin{proof}

Let $\tX_{\textrm{base}}$ be the samples captured by $e^{\textrm{base}}$. The objective can be written as 
\begin{equation*}
    \min_{e^{\textrm{add}} \subseteq C_{q}} \supp(\{e^{\textrm{base}}, e^{\textrm{add}}\}, \tX) 
    = \min_{e^{\textrm{add}} \subseteq C_{q}} \left| S_{e^{\textrm{base}}}(\tX) \cap S_{e^{\textrm{add}}}(\tX)\right|
    = \min_{e^{\textrm{add}} \subseteq C_{q}} \left| S_{e^{\textrm{add}}}(\tX_{\textrm{base}})\right|.
\end{equation*}

Minimization of the intersection is equivalent to maximizing the complement, i.e, $$\max_{e^{\textrm{add}} \subseteq C_{q}} \left| \left( S_{e^{\textrm{add}}}(\tX_{\textrm{base}})\right)^c \right| = \max_{e^{\textrm{add}} \subseteq C_{q}} \left| \left( \cap_{c_j \in e^{\textrm{add}}}S_{c_j}(\tX_{\textrm{base}})\right)^c \right| $$
which, by De Morgan's law, $$=
\max_{e^{\textrm{add}} \subseteq C_{q}} \left| \left( \cup_{c_j \in e^{\textrm{add}}} S_{c_j}(\tX_\textrm{base})^c\right) \right|. $$

Since $c_j$ are binary conditions, the complement of samples captured by $c_j$ is the collection of samples captured by $\neg c_j$, i.e., 
$$\max_{e^{\textrm{add}} \subseteq C_{q}} \left| \left( \cup_{c_j \in e^{\textrm{add}}} S_{\neg c_j}(\tX_{\textrm{base}}) \right) \right|.$$
Therefore, Problem \eqref{eq:prob2} is the same as the maximum coverage problem. 
\end{proof}


\vfill

\section{Datasets and experimental setup}\label{app:setup}
We use three credit datasets FICO credit risk dataset \citep{competition}, German credit dataset from UCI \citep{Dua:2019}, and a loan approval prediction problem dataset from Kaggle \citep{kaggle}. Details about these datasets are in Table \ref{tab:dataset}. 

\begin{table}[h]
    \centering
    \begin{tabular}{|l|c|c|l|}\hline
       Dataset &  Samples & Features & Description \\ \hline
       FICO  &  10459 & 23 & whether someone would default on a loan\\\hline
       German credit & 1000 & 20 & predict the approval of credit applications \\\hline
       Kaggle loan prediction problem & 480 & 11 & predict the approval of loan applications\\\hline
       
    \end{tabular}
    \vspace{1mm}
    \caption{Dataset summary}
    \label{tab:dataset}
\end{table}

All experiments are run on a 2.7Ghz (768GB RAM 48 cores) Intel Xeon Gold 6226 processor. Cplex Studio 22.1 is used for FaithfulDefense IP and FaithfulDefense IP-RA. The time limit is set to 180 seconds. To train the surrogate model, we use CART, Random Forest, and GBDT from scikit-learn 
\citep{scikit-learn}. We set CART's max\_depth to 5 and use the default setting for other hyperparameters. 

\section{More experimental results}\label{app:more_results}

In this section, we show more experimental results. 


\textbf{Interpretable models are as accurate as complex models. } We compare models produced by FastSRS, Random Forest, and GBDT on the original datasets in Table \ref{tab:base_model_vs_black_box}. FastSRS can achieve performance comparable to complex models, indicating that interpretable models can replace complex models in real applications. This motivates us to develop models that are interpretable but not transparent for practical use. 
\begin{table}[h]
    \centering
    \begin{tabular}{|l|c|c|c|l|}\hline
       Dataset & FastSRS & Random Forest & GBDT \\ \hline
       FICO  & 0.713  & 0.725 & \textbf{0.73} \\\hline
       German credit & 0.755  & \textbf{0.765} & 0.755 \\\hline
       Kaggle loan prediction problem & \textbf{0.844}  & 0.802 & 0.833\\\hline
       
    \end{tabular}
    \vspace{1mm}
    \caption{Test accuracy of interpretable models versus complex models on the original datasets. FastSRS achieves test accuracy comparable to Random Forest and GBDT. }
    \label{tab:base_model_vs_black_box}
\end{table}

\begin{figure}[h]
    \centering
    \includegraphics[width=0.8\textwidth]{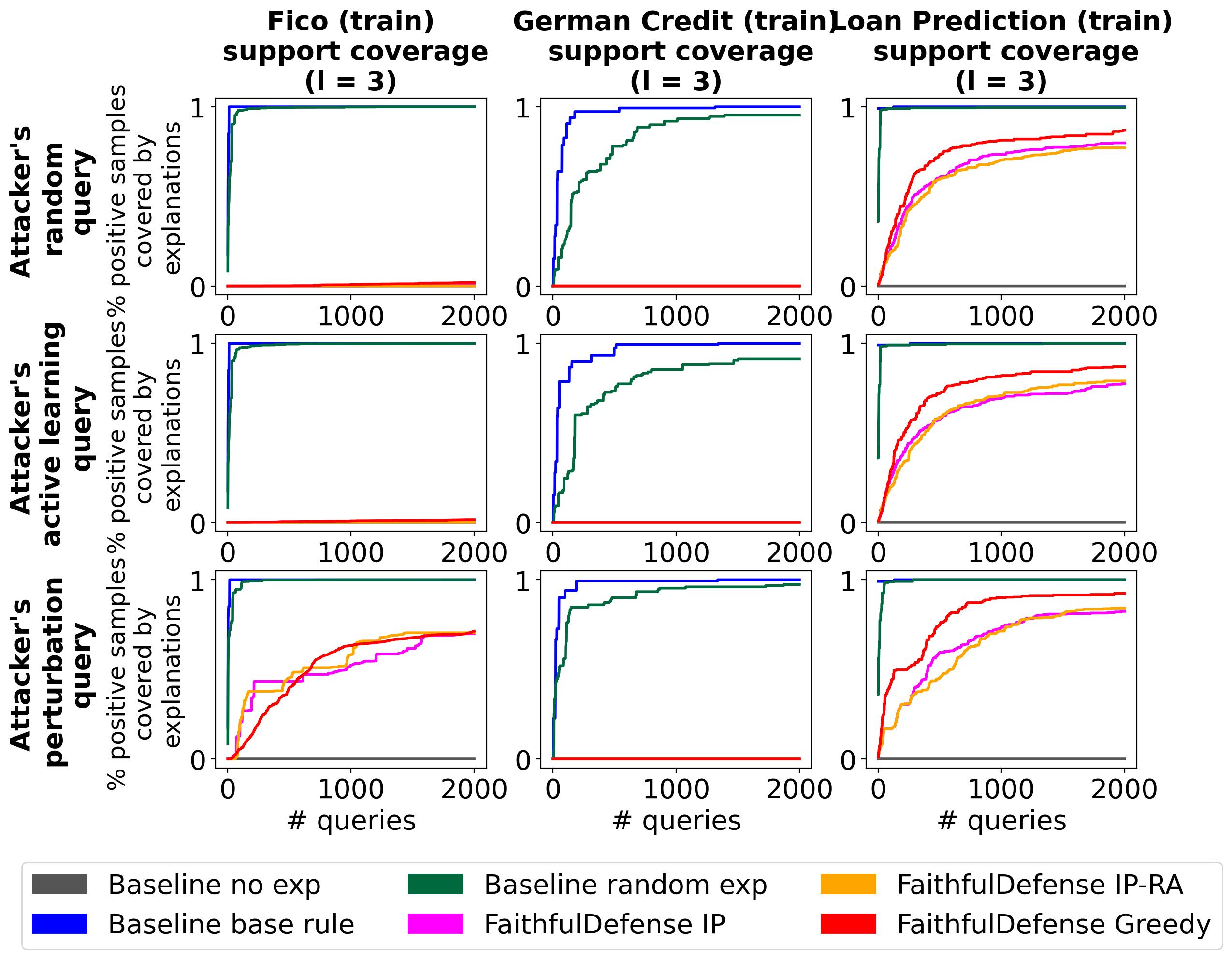}
    \caption{Number of queries vs. the proportion of positive samples covered by explanations on the training set. (max length $l=3$).}
    \label{fig:supp_train}
\end{figure}

\textbf{How Many Positive Training Points are Covered by the Explanation.} Figure \ref{fig:supp_train} demonstrates that our FaithfulDefense (red, orange, and pink curves) reveals less information (i.e., captures fewer positive samples) in training sets for all three datasets using three different querying strategies, as the red, orange, and pink curves consistently fall below the green and blue curves. 

\begin{figure}[h]
\centering
\includegraphics[width=0.9\textwidth]{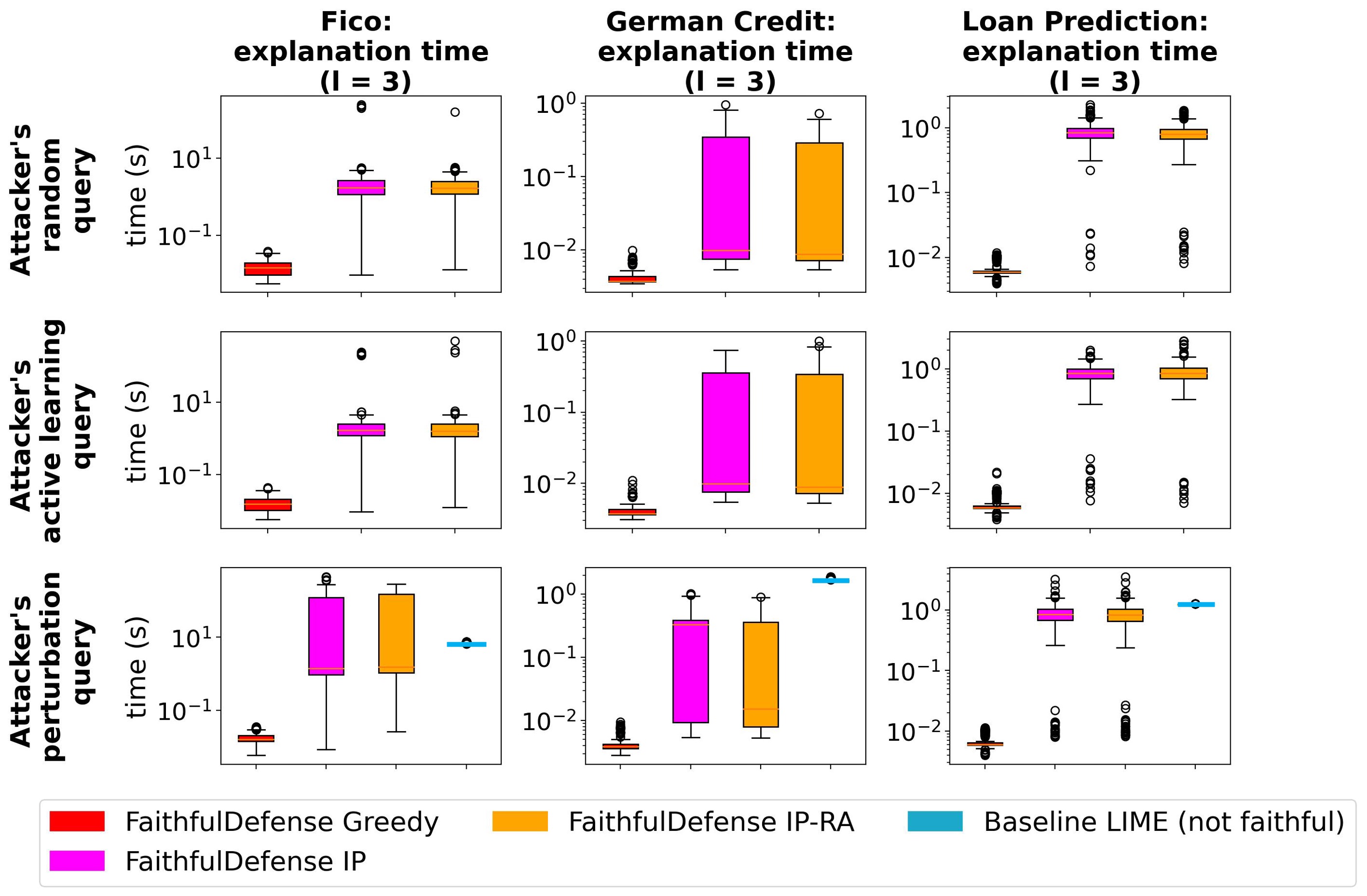}
\caption{Time consumption of generating explanations. (max length $l=3$)}
\label{fig:exp_time_app}
\end{figure}

\textbf{Explanation Generation Timing.} 
Figure \ref{fig:exp_time_app} shows the explanation generation time on three datasets when three different querying strategies are used. 
We observe that the time taken by our FaithfulDefense Greedy is generally fast, usually less than 0.1 second for different querying strategies, while FaithfulDefense IP and FaithfulDefense IP-RA have higher time costs.

\begin{figure}
\centering
\includegraphics[width=0.8\linewidth]{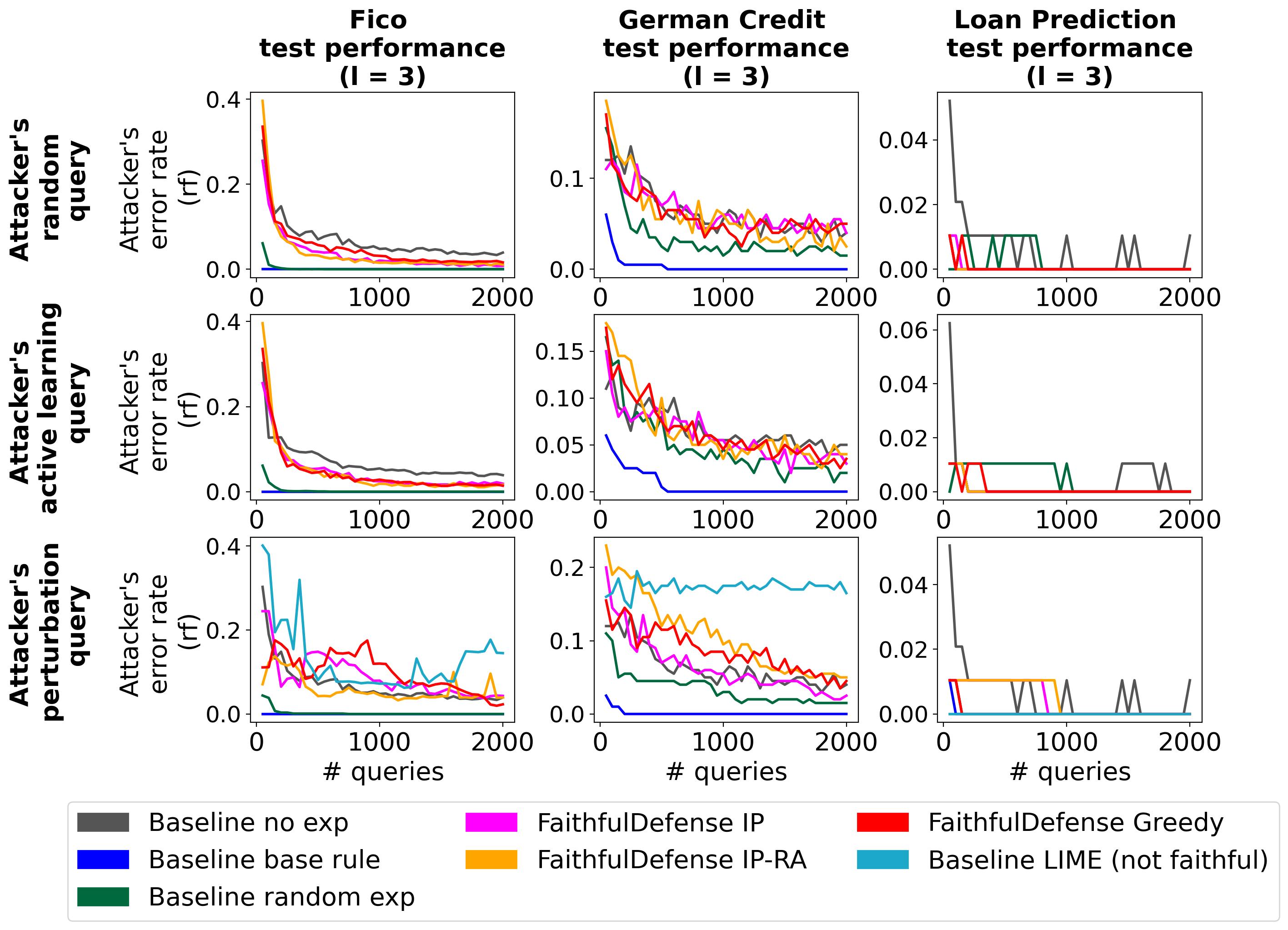}
\includegraphics[width=0.8\linewidth]{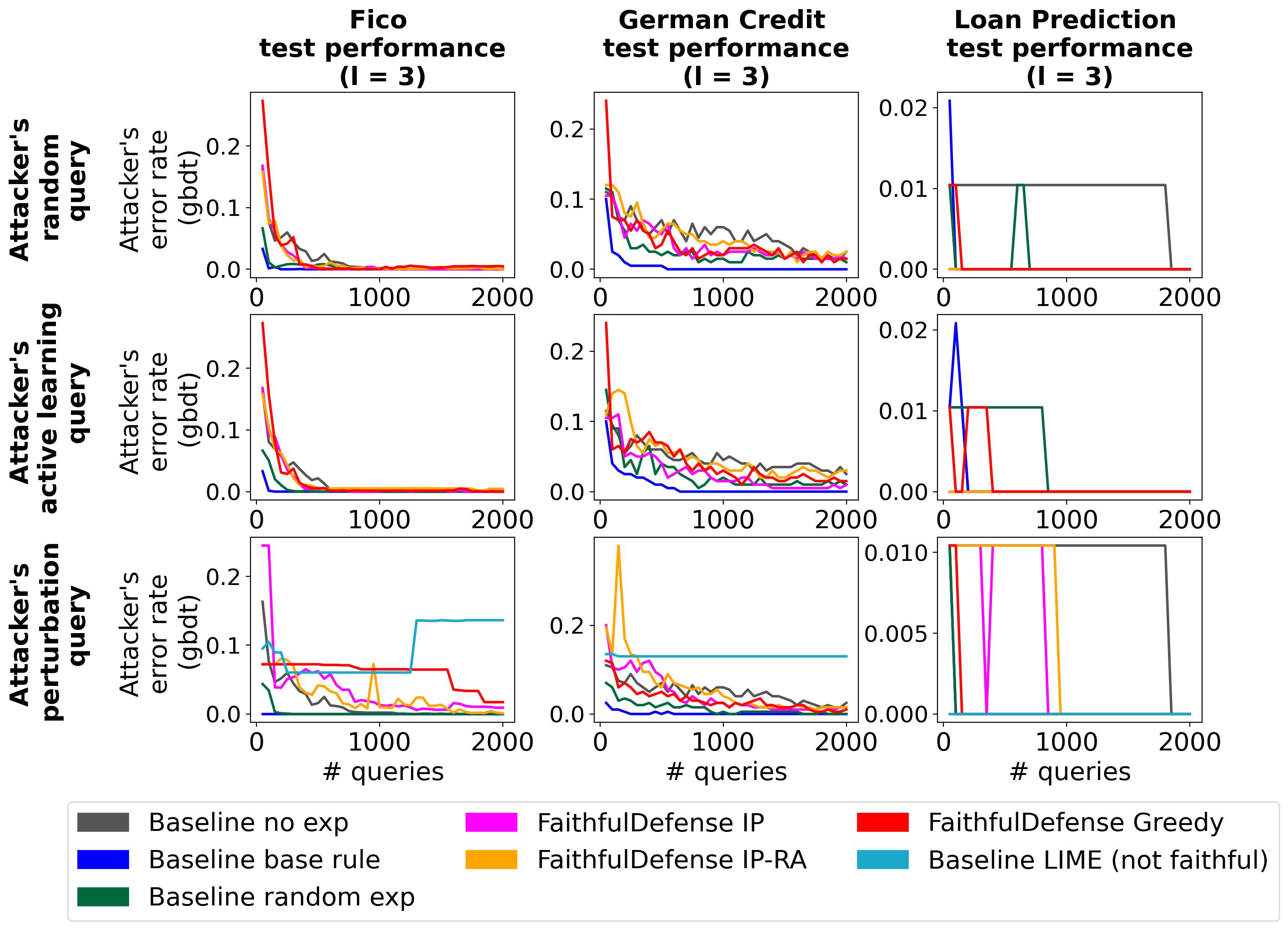}
\caption{Comparison of test performance between base and surrogate models. Random Forest (top) and GBDT (bottom) are used by the attacker to train the surrogate model. (max length $l=3$).}
\label{fig:rf_gbdt_surrogate}
\end{figure}

\textbf{Test Accuracy of Surrogate Models.} 
Due to the space limitation, we only showed the test performance of surrogate models trained using CART in the main paper. Figure \ref{fig:rf_gbdt_surrogate} shows the test performance when Random Forest and GBDT are used to train the surrogate model. We use the default setting for these two model configurations. Test performance using Random Forest is similar to that of CART, where providing our explanations usually requires more queries compared with the other two baselines and sometimes providing our explanations yields test performance close to providing no explanations. GBDT is more powerful than CART and Random Forest; however, Random Forest and GBDT are black box models. They cannot be used to replace the interpretable models to return faithful explanations.

\begin{figure}
    \centering
    \includegraphics[width=.7\textwidth]{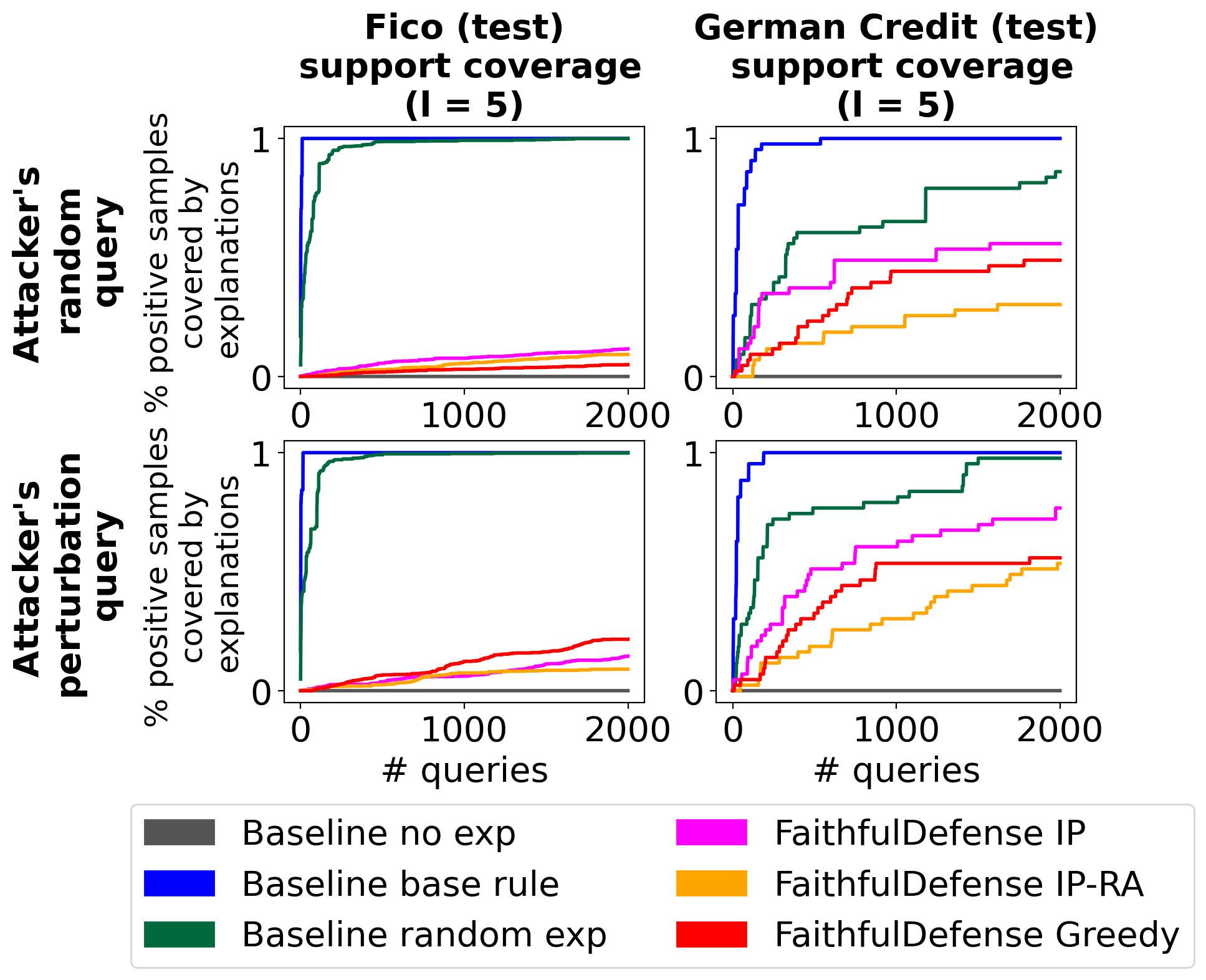}
    \includegraphics[width=.7\textwidth]{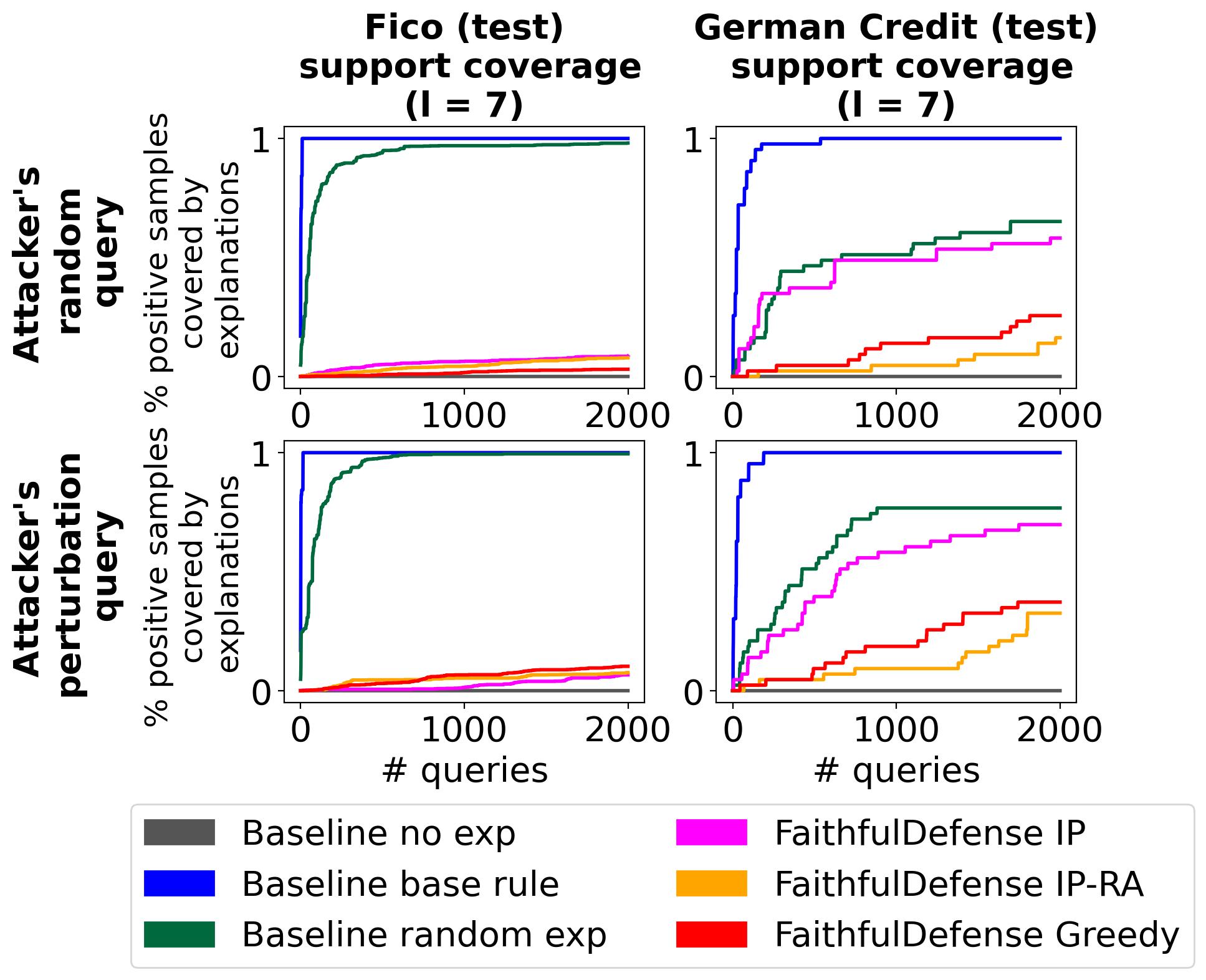}
    \caption{Number of queries vs$.$ the proportion of positive samples covered by explanations on the test set (max length $l =5$ and $l = 7$).}
    \label{fig:supp_5_7}
\end{figure}

\begin{figure}
    \centering
        \includegraphics[width=.7\textwidth]{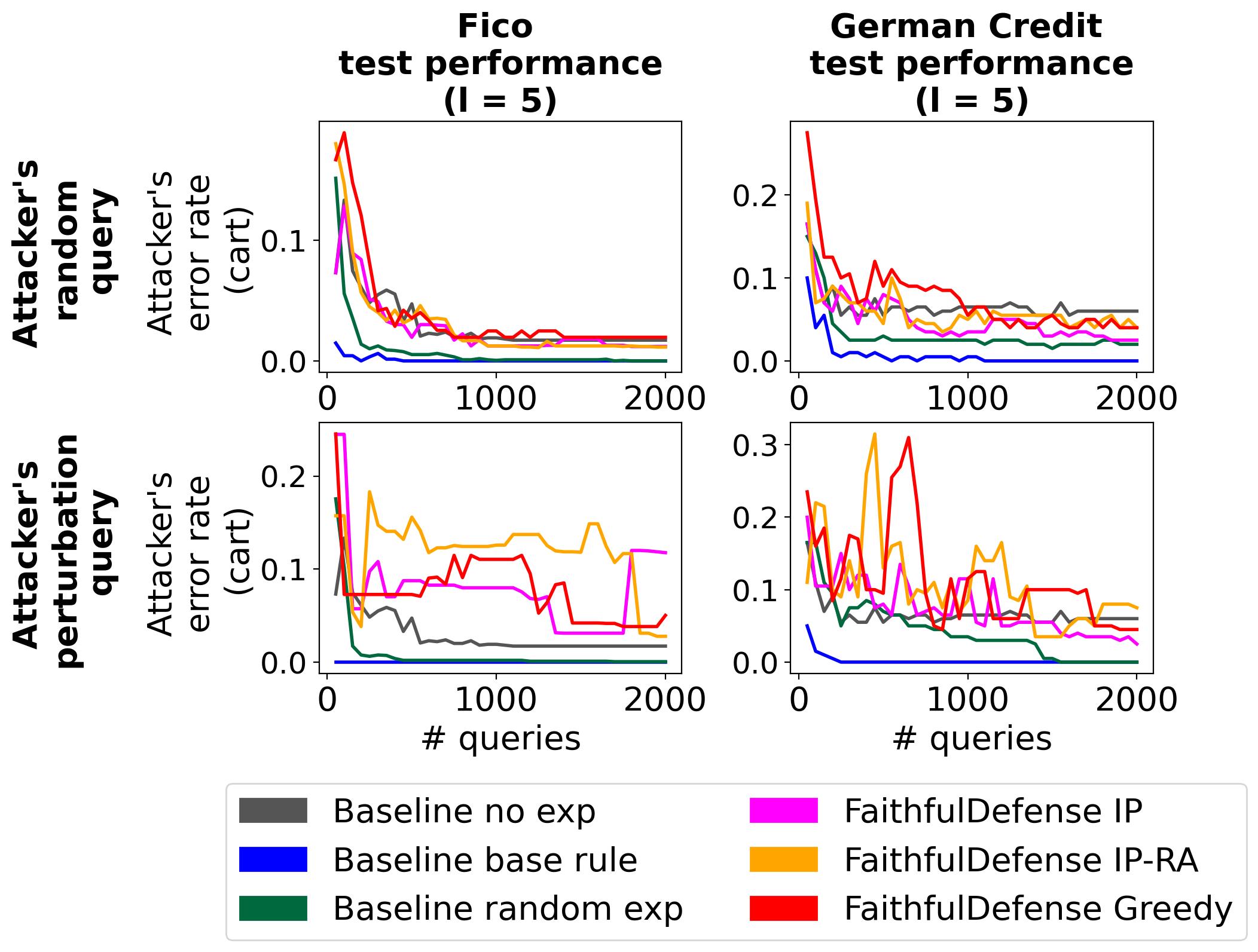}
        \includegraphics[width=.7\textwidth]{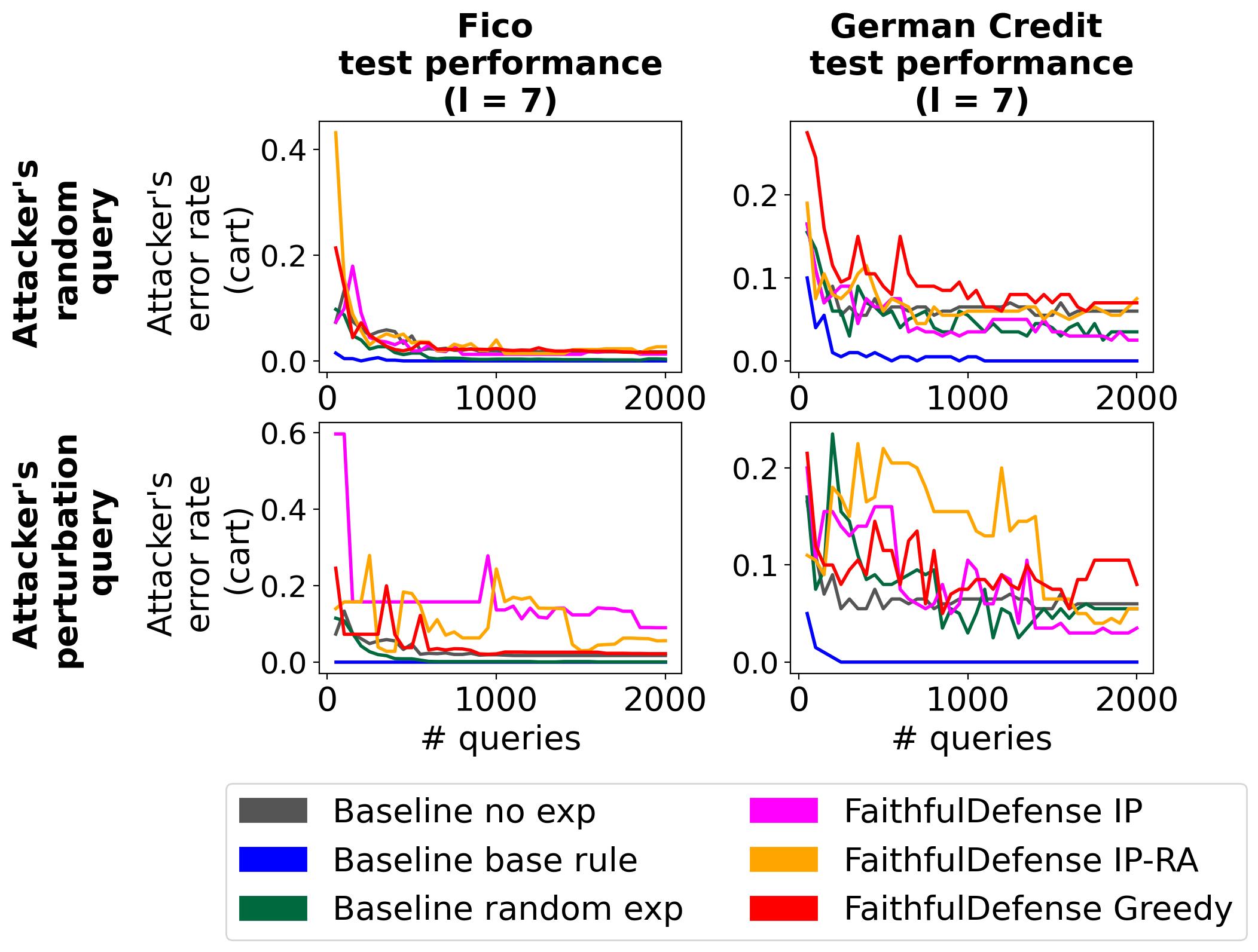}    
    \caption{Comparison of test performance when max length $l=5$ and $l=7$ with respect to the CART model.}
    \label{fig:accu_5_7_cart}
\end{figure}

\begin{figure}
    \centering
        \includegraphics[width=.7\textwidth]{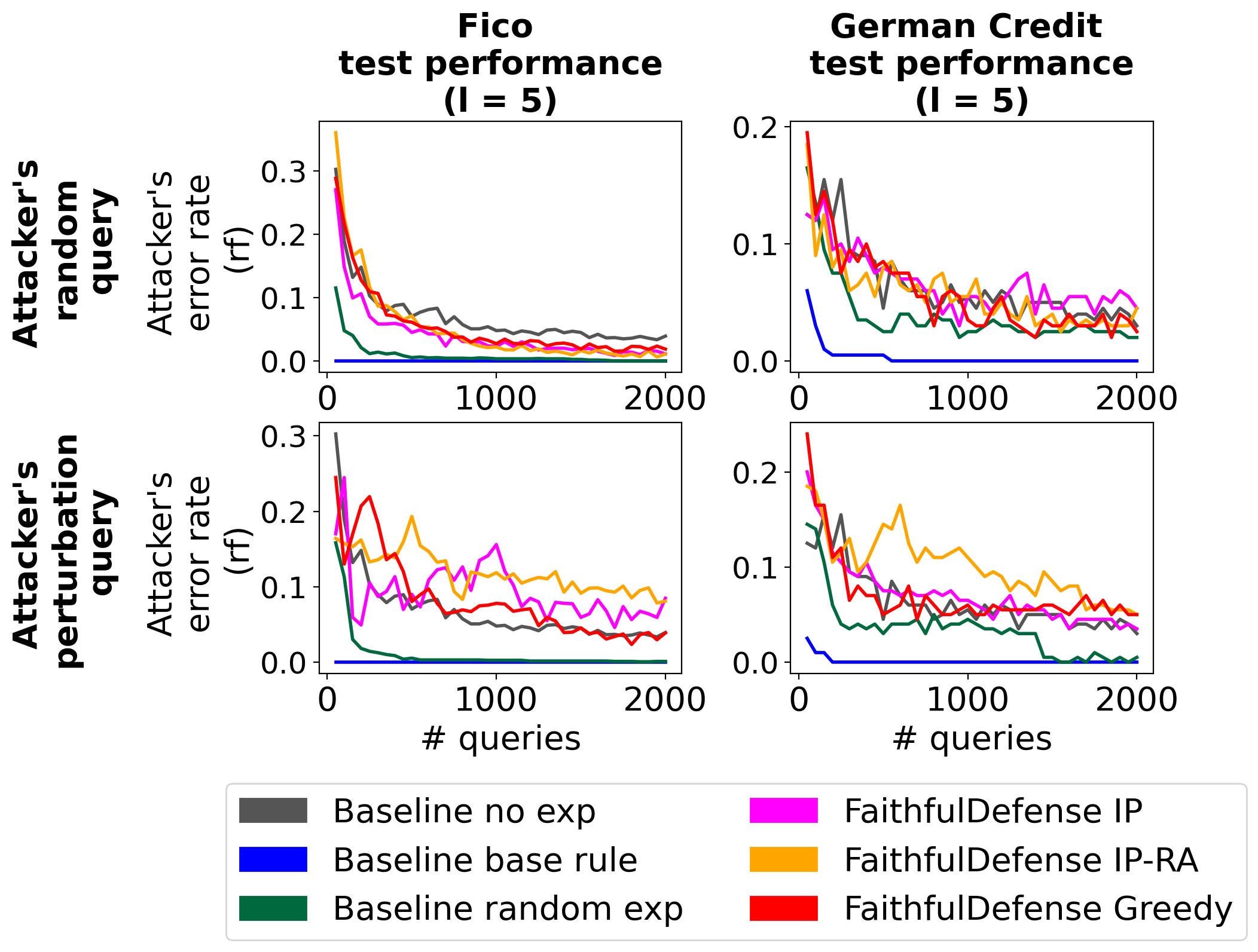}
        \includegraphics[width=.7\textwidth]{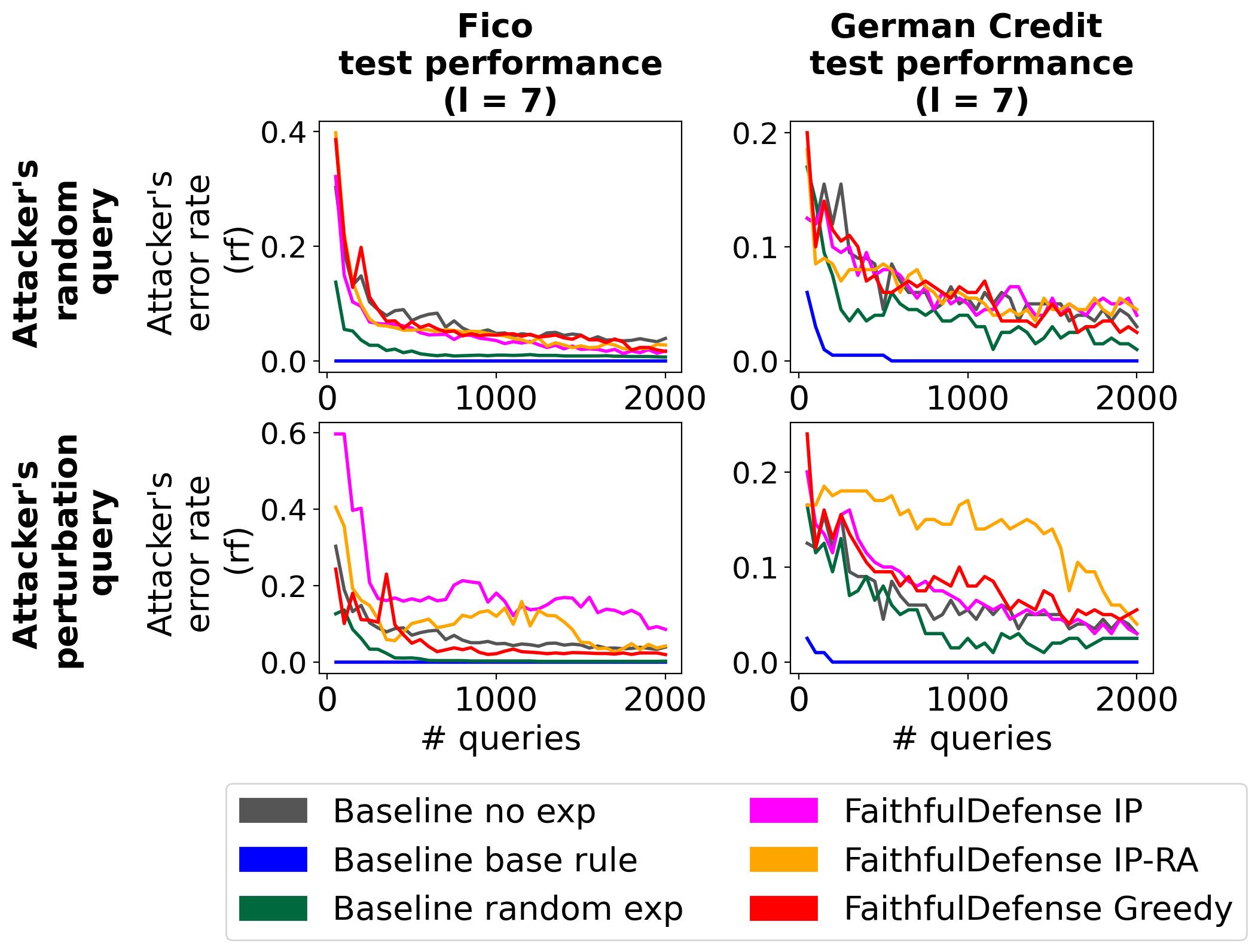}    
    \caption{Comparison of test performance when max length $l=5$ and $l=7$ with respect to the Random Forest model.}
    \label{fig:accu_5_7_rf}
\end{figure}

\begin{figure}
    \centering
        \includegraphics[width=.7\textwidth]{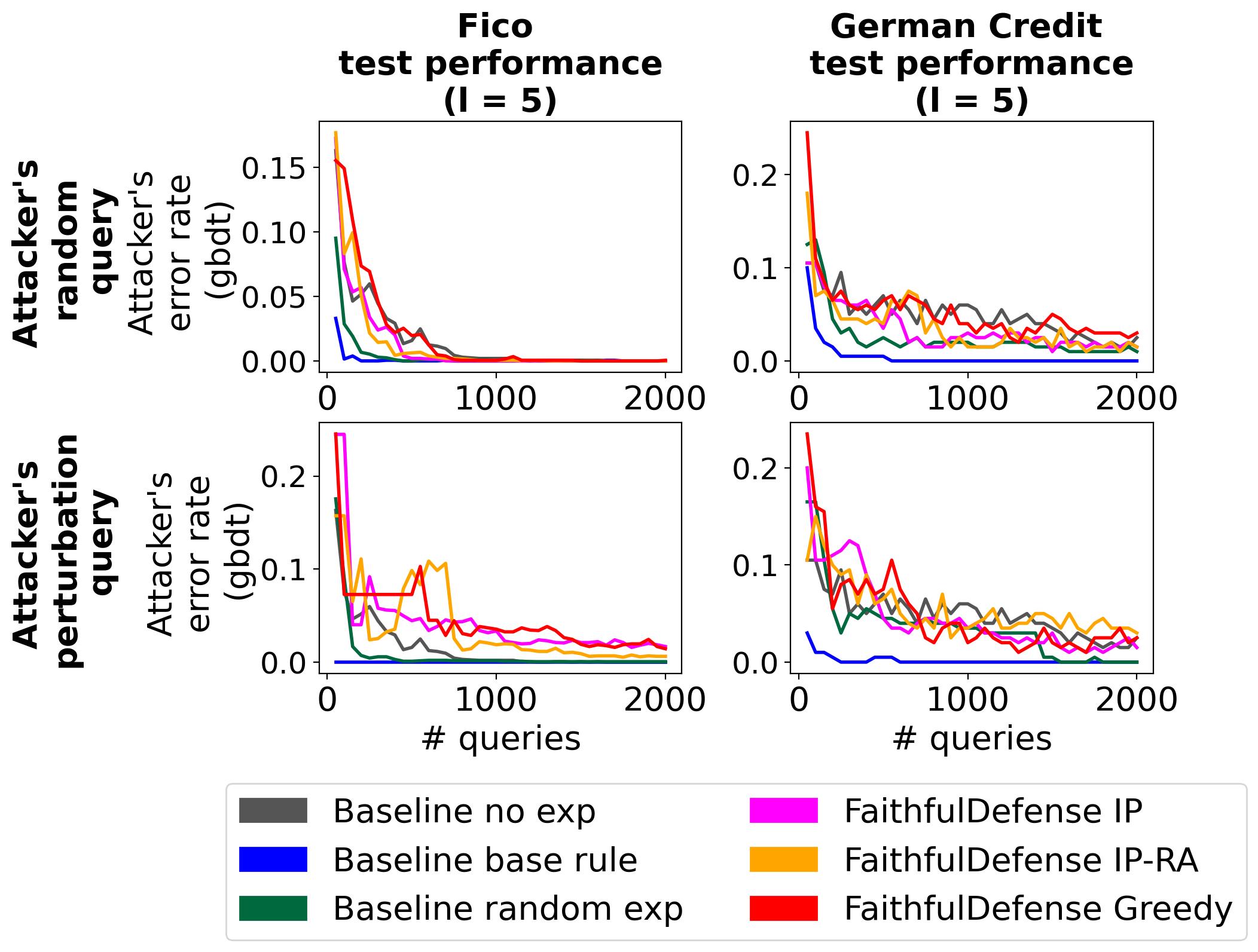}
        \includegraphics[width=.7\textwidth]{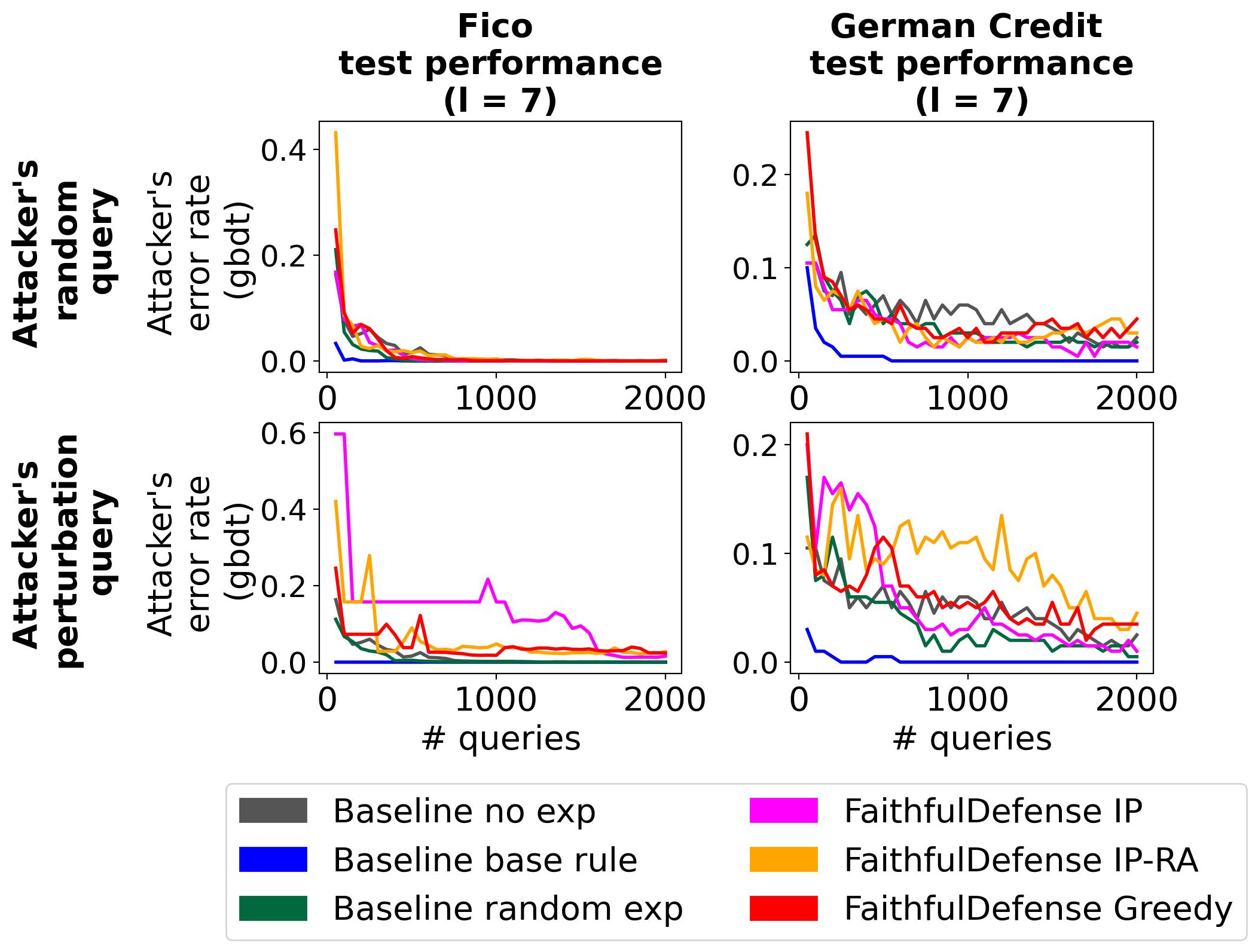}    
    \caption{Comparison of test performance when max length $l=5$ and $l=7$ with respect to the GBDT model.}
    \label{fig:accu_5_7_gbdt}
\end{figure}

\textbf{Impact of Changing Max Length.} We also study how the value of max length $l$ influences information leakage and the attacker's performance. Figure \ref{fig:supp_5_7} shows the number of queries versus support coverage when $l$ is set to 5 and 7. For all explanation methods, support coverage decreases as $l$ increases on the test set since more conditions are used in the explanation. Our FaithfulDefense outperforms the baselines, as the red, orange, and pink curves are always below the green and blue curves. Additionally, our FaithfulDefense outperforms the baselines on surrogate model performance. As shown in Figure \ref{fig:accu_5_7_cart}-\ref{fig:accu_5_7_gbdt}, the red, orange, and pink curves for the FICO and German credit datasets are generally above the green and blue curves, regardless of the attacker's querying strategy or the model class used to train the surrogate model. 

\newpage
\section{FaithfulDefense for other model classes}\label{ref:app:gams}

Generalized additive models \citep{hastie1990generalized} linearly combine flexible component functions for each feature:
\begin{equation}
    g(E[y]) = \omega_0 + f_1(x_1) + f_2(x_2) + \cdots + f_p(x_p),
\end{equation}
where $x_j$ indicates the $j$th feature, the $f_j$'s are learned univariate component functions that are possibly nonlinear, and $g(\cdot)$ is a link function, e.g., the identity function for regression or the inverse logistic function for classification. Each shape function $f_j$ operates on only one feature $x_j$, thus the shape functions can directly be plotted. This makes GAMs interpretable since the entire model can be easily visualized.

In practice, each continuous feature is usually divided into bins \citep{lou2012intelligible, liu2022fast}, thereby its shape function can be viewed as a set of step functions, i.e.,
\begin{equation}
    f_j(x_j) = \sum_{k=0}^{B_j-1}\omega_{j,k}\cdot \mathbf{1}[b_{j,k}<x_j \leq b_{j,k+1}],
\end{equation}
where $\{b_{j,k}\}_{k=0}^{B_j}$ are the bin edges of feature $j$, leading to  $B_j$ total bins. 

Sparsity regularization, i.e., the $\ell_0$ penalty on the number of steps, is often used to encourage generalization and avoid constructing overly complicated models. A sparse GAM model with piecewise constant shape functions can be efficiently converted into logic models. For example, given $p$ shape functions and each shape function has $B_j$ bins, the GAM model can be converted into a multi-split decision tree with at most depth $p$. In reality, not all leaves have to reach the depth $p$. If concatenating bins from other shape functions will not change the leaf prediction, then we can stop early. Any decision tree can then be converted into decision sets by extracting leaf paths with positive predictions. 


\section{Relationship to recourse}
Our proposed FaithfulDefense does not aim to provide recourse. There are three issues in providing recourse discussed below: (1) Recourse has technical challenges in being too prescriptive; (2) Recourse reveals a tremendous amount about the model, making it difficult to keep it non-transparent; (3) Laws require explanations, but not recourse. This could easily be due to the two issues listed above. Let us discuss this in more depth below.

(1) Recourse technical challenges and being too prescriptive. Recourse requires being able to change the features, and knowing costs for each possible change to the features. 

- Many features may not be able to be changed. For loan decisions, the features are typically based on credit history and job status, and there is generally no way for loan applicants to change those. Thus, recourse may simply not be possible.

- In other situations where features could be changed, there is a cost for changing each feature in the recourse framework, but those costs might be unknown or too high. How much would it cost the loan applicant to change jobs to make more per month? It is not clear they would be able to do it at all let alone having the bank know the cost for the applicant to do that. Thus, any possible recourse given to the user may not estimate these costs correctly and may not be actionable in reality (i.e., may be too prescriptive).

(2) Recourse reveals a lot about the model. If we told the user that changing feature 1 would change the decision, they would then know that feature 1 is actually used in the model, whereas our approach hides that information in order to keep the model non-transparent. So, any defensive algorithm would fail to keep the model's variables as a secret. Since the bank would not use a model that is easily revealed, it most likely would resort to a black box with unfaithful explanations again. Another option is to provide an unfaithful recourse, but this would defeat the point of providing the recourse in the first place.

(3) The obligation of loan lenders to give an explanation for each denial is sourced from ``Right to explanation'' and other similar laws (e.g., in the Code of Federal Regulations, the \href{https://www.ecfr.gov/current/title-12/part-1002/section-1002.9#p-1002.9(b)(2)}{``Form of Equal Credit Opportunity Act (ECOA) notice and statement of specific reasons''}). Those laws only govern the right to explanation, not the right to recourse, possibly due to the issues mentioned above in recourse not actually being possible, and, if possible, being impractical to compute effectively.


\end{document}